\documentclass[letterpaper, 10 pt, conference]{ieeeconf}  

\IEEEoverridecommandlockouts                              

\usepackage{graphicx} 
\usepackage{epsfig} 
\usepackage{mathptmx} 
\usepackage{times} 
\usepackage{amsmath} 
\usepackage{amssymb}  
\usepackage{booktabs}
\usepackage{subcaption}
\newcommand{\etal}{\textit{et al.}}
\usepackage{tikz}
\usepackage{eso-pic} 

\title{\LARGE \bf
Cal or No Cal? - Real-Time Miscalibration Detection of LiDAR and Camera Sensors}

\author{Ilir Tahiraj$^{*}$,  Jeremialie Swadiryus, Felix Fent, Markus Lienkamp
\thanks{$^{*}$ Corresponding author. {\tt\small ilir.tahiraj@tum.de}.}
\thanks{Authors are with the TUM School of Engineering and Design, Chair of Automotive Technology,
        Technical University of Munich.}%
}

\newcommand\copyrighttext{%
    \footnotesize \textcopyright{} This work has been submitted to the IEEE for possible publication. Copyright may be transferred without notice, after which this version may no longer be accessible.
}
\newcommand\copyrightnotice{%
    \begin{tikzpicture}[remember picture, overlay]
        \node[anchor=south, yshift=10pt] at (current page.south) {\fbox{\parbox{\dimexpr\textwidth-\fboxsep-\fboxrule\relax}{\copyrighttext}}};
    \end{tikzpicture}
}

\begin{document}

\maketitle
\copyrightnotice
\thispagestyle{empty}
\pagestyle{empty}

\begin{abstract}
The goal of extrinsic calibration is the alignment of sensor data to ensure an accurate representation of the surroundings and enable sensor fusion applications. From a safety perspective, sensor calibration is a key enabler of autonomous driving. In the current state of the art, a trend from target-based offline calibration towards targetless online calibration can be observed. However, online calibration is subject to strict real-time and resource constraints which are not met by state-of-the-art methods.  This is mainly due to the high number of parameters to estimate, the reliance on geometric features, or the dependence on specific vehicle maneuvers. To meet these requirements and ensure the vehicle's safety at any time, we propose a miscalibration detection framework that shifts the focus from the direct regression of calibration parameters to a binary classification of the calibration state, i.e., calibrated or miscalibrated. Therefore, we propose a contrastive learning approach that compares embedded features in a latent space to classify the calibration state of two different sensor modalities. Moreover, we provide a comprehensive analysis of the feature embeddings and challenging calibration errors that highlight the performance of our approach. As a result, our method outperforms the current state-of-the-art in terms of detection performance, inference time, and resource demand. The code is open source and available on https://github.com/TUMFTM/MiscalibrationDetection.
\end{abstract}

\section{Introduction}
\label{sec:intro}

Ensuring safety is the primary challenge in the development of autonomous driving systems. One key aspect of it is the development of an accurate environment model, which helps autonomous vehicles (AVs) to understand and predict their surroundings by fusing data from camera and LiDAR sensors~\cite{Wang2023, Yu2023, Dong2023}. Active research continues to improve sensor fusion and 3D object detection, with efforts focused not only on enhancing the accuracy but also on making sensors more reliable in challenging conditions such as adverse weather conditions, sensor failures, spatial and temporal misalignment, and feature sparsity~\cite{Bai2022, Xie2023, Song2023, Liu2023}. 

While sensor fusion is an enabler for reliable scene understanding, it also introduces new challenges. In particular, sensor synchronization and calibration. Sensor fusion requires aligned sensor data to ensure the accuracy and reliability of object detection algorithms. Therefore, accurate sensor calibration is critical to the safety of AVs.

\begin{figure}
    \centering
    \includegraphics[width=1\linewidth]{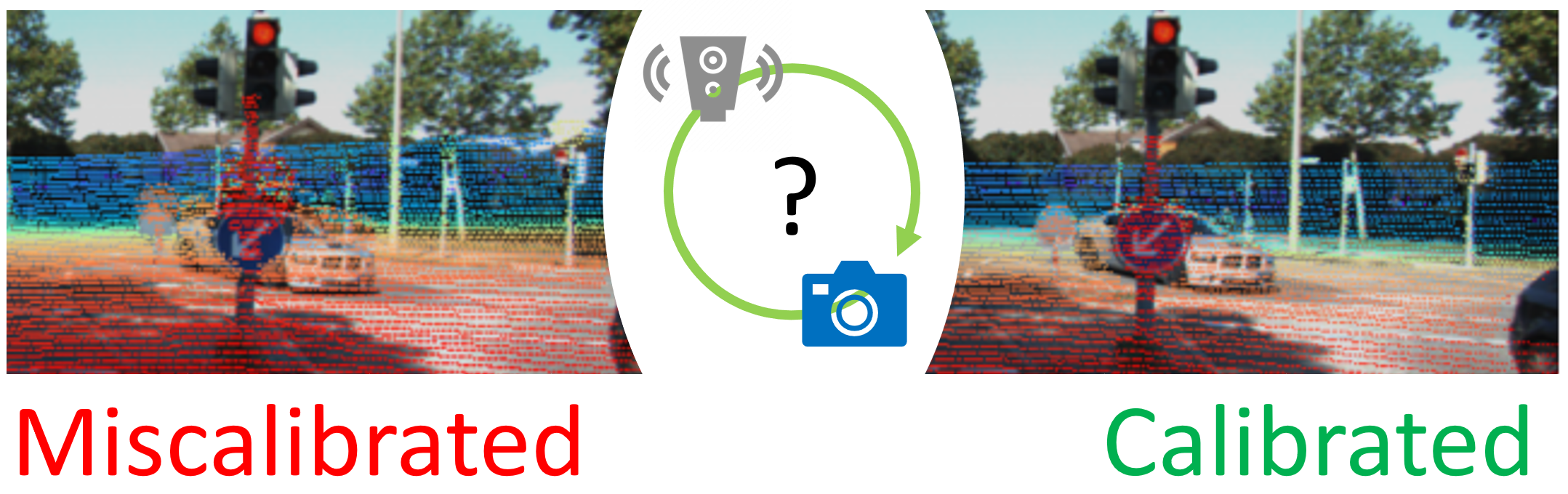}
    \caption{Our miscalibration detection framework takes RGB images and LiDAR point clouds as input. The point clouds are projected onto the image and used to detect miscalibration between the sensors.}
    \label{fig:intro}
    \vspace{-12pt}
\end{figure}

Sensor calibration, and more specifically extrinsic sensor calibration, is the process of obtaining the rigid transformations between sensor coordinate systems. To address this problem, two strategies are used in autonomous driving: Calibration before or after the deployment of the autonomous vehicle. The former concept can be performed using specific calibration targets in a dedicated environment such as a vehicle factory. The latter involves the use of online calibration methods that aim to (re-)calibrate the sensors in an unstructured environment during vehicle operation. The second approach is mainly motivated by the fact that sensor miscalibration is caused by temperature changes and/or vehicle vibrations that occur after vehicle deployment.

Despite considerable efforts to develop online calibration methods, several challenges remain~\cite{survey}. First, identifying common features in multimodal sensor data and open scenes remains a complex task, preventing online calibration methods from achieving the accuracy of offline target-based approaches. Second, online calibration — particularly learning-based approaches (see Section~\ref{sec:regressionbased}) — still suffers from limited generalization in real-world scenarios.

Online sensor calibration traditionally involves the regression of at least 12 parameters for each individual extrinsic sensor-to-sensor calibration, namely the components of the rotation matrix and the translation vector. The number of parameters to regress increases even further when estimating more than two sensor calibrations or the intrinsics simultaneously. Therefore, regression-based calibration algorithms are resource intensive~\cite{LCCNet, CalibDepth} and difficult to verify for real-world, safety-critical systems~\cite{survey}. Furthermore, many online calibration algorithms require certain environmental conditions to be met, such as the presence of distinctive geometric features or the execution of specific driving maneuvers~\cite{Yuan2021, PBACalib}. These requirements effectively limit the capacity of online calibration algorithms to ensure anytime-safety of the system. 
In addition, online recalibration algorithms are mainly trained to regress the extrinsic parameters. Intrinsic calibration errors, such as focal length or principal point offsets, can lead to incorrect recalibration of the extrinsic parameters because such intrinsic errors cannot be easily distinguished from extrinsic errors. 

In response to these challenges, we provide a continuous sensor monitoring approach and propose shifting the focus from a regression to a classification-based framework. By formulating the online calibration problem as a classification task, we can simplify the model while improving both efficiency and reliability. Our classification-based framework allows the identification of binary calibration states rather than the regression of extrinsic parameters, enabling the system to detect and respond to miscalibrations in real time with less computational effort compared to a full recalibration. Our main contributions are as follows:
\begin{itemize}
    \item We introduce an open-source framework for miscalibration detection using a self-supervised learning architecture and feature-based classification.
    \item This is the first two-stage learning approach for miscalibration detection, where one stage learns robust representations of the miscalibrated sensor data. The second stage uses these representations to train the classification task.
    \item We are the first to provide a framework that shows that it can robustly detect a miscalibration in the presence of intrinsic errors.
    \item Using Centered Kernel Alignment (CKA), we analyze feature representations, enabling a simple architecture with faster inference and reduced computational needs suited for real-time use.
    \item Our approach achieves state-of-the-art detection, with inference time 6$\times$ faster and a model size 42\% smaller than existing methods.
\end{itemize}

\section{Related Work}
\label{sec:relatedwork}
Current state-of-the-art LiDAR-to-camera sensor calibration methods estimate transformation parameters by identifying correspondences and optimizing the alignment between LiDAR points and camera images. These methods often use appearance-based optimizations or deep learning techniques to achieve the desired calibration accuracy. We categorize these methods as regression-based approaches, focusing on those that perform online calibration to address on-site sensor misalignment. In contrast, we categorize miscalibration detection methods as classification-based tasks that aim to identify whether two or more sensors are properly aligned.
 
\subsection{Regression-based Concepts}\label{sec:regressionbased}
LiDAR-to-camera calibration methods mainly build correspondences from LiDAR point clouds to images or vice versa. These correspondence-based methods can be further divided into explicit and learned correspondence search~\cite{survey}.  
\subsubsection*{Explicit Correspondences}
Examples of approaches that match explicit correspondences using geometric features are presented in~\cite{Yuan2021, PBACalib, CRLF}. Yuan~\etal~\cite{Yuan2021} rely on finding edge correspondences to perform LiDAR-to-camera calibration. They first extract edges from 3D point clouds and estimate the transformation matrix by aligning them with edges from 2D images. In~\cite{PBACalib}, distinctive planar features are captured from both the LiDAR and the camera data to estimate the extrinsic calibration matrix via plane-constrained bundle adjustment. CRLF~\cite{CRLF} performs a two-step calibration process by using line features extracted from the image and LiDAR point clouds. First, an initial calibration matrix is estimated by line fitting in both LiDAR and camera space, before a refinement step estimates the final transformation.

Recent approaches aim to perform online calibration using non-geometric features. The approach in~\cite{Ishikawa2024} takes advantage of the fact that the projected intensities (LiDAR-to-camera) have a high variance when there are errors in the calibration parameters and formulates an optimization problem based on the intensity variance. The calibration algorithms presented in~\cite{SemAlign, Luo2024, SGCalib} rely on semantic information extracted from the sensor modalities, i.e. correspondences exist between point clouds and images based on the same semantic information.

\subsubsection*{Learned Correspondences}
\begin{figure}
    \centering
    \includegraphics[width=1\linewidth]{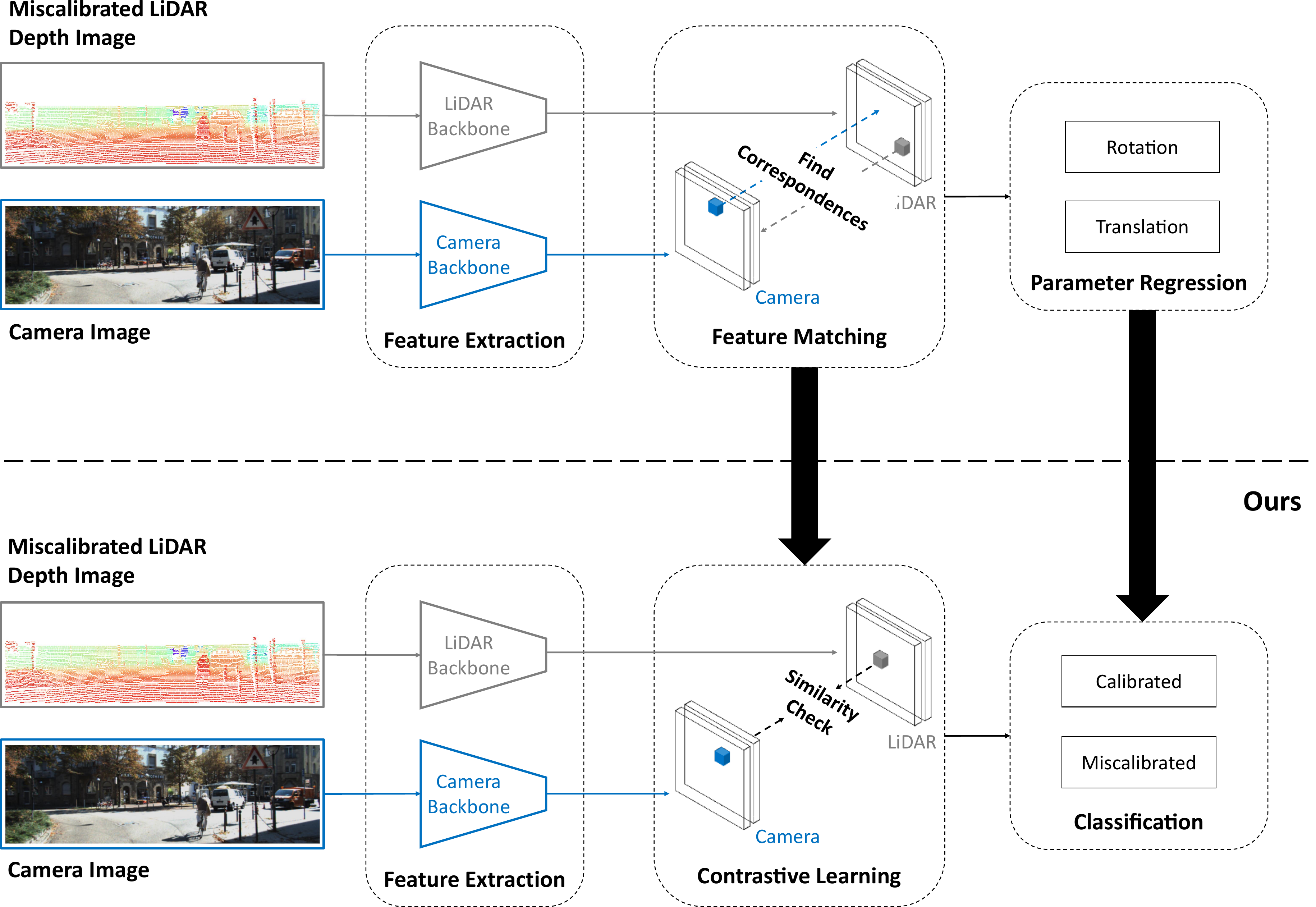}
    \caption{\textbf{Above}: The general approach for online estimation of calibration parameters. The feature matching layer serves the correspondence finding process. The final layer regresses the translation and rotation of the extrinsic calibration transforms.
    \textbf{Below}: Our approach aims to learn representations that describe the similarity or dissimilarity between the multimodal features. The final layer classifies whether the inputs are calibrated or miscalibrated.}
    \label{fig:our_architecture}
    \vspace{-12pt}
\end{figure}

Methods with learned correspondences have proven to be powerful for use in extrinsic sensor calibration. These methods use deep learning techniques to find correspondences between the LiDAR and camera domain and estimate the calibration matrix in an end-to-end fashion~\cite{Schneider2017, CalibNet, RGGNet, FusionNet}. Schneider~\etal~\cite{Schneider2017} introduced RegNet, a LiDAR-to-camera calibration framework with an end-to-end approach: a convolutional neural network that performs feature extraction, correspondence matching, and the regression of a 6 DoF calibration matrix. CalibNet~\cite{CalibNet} proposes a more general approach, which is independent of camera characteristics and sensor configurations. In general, these methods are based on the architectures shown in Fig.~\ref{fig:our_architecture}.

Wang~\etal~\cite{FusionNet}, similar to traditional techniques in extrinsic sensor calibration, incorporate the coarse-to-fine calibration procedure into a learning-based framework. Hu~\etal~\cite{Hu2024} follow the same idea of iterative refinement and introduce HIFMNet, which constructs different cost volumes, one to initialize with a coarse extrinsic calibration to find large deviations between the depth maps and the RGB images and another for an iterative refinement of the extrinsics. 

SOAC~\cite{SOAC} propose to use Neural Radiance Fields (NeRF) for extrinsic and temporal sensor calibration. It is trained with observations from different sensors to generate a 3D scene representation and optimize the extrinsic calibration parameters to fit this representation. The work in~\cite{CalibDepth} assumes that matching correspondences within the same data representation is most effective. Therefore, images and point clouds are transformed into depth maps, which are fed into a feature extraction, feature matching, and parameter regression module.

While methods based on explicit correspondences generalize well, they require a good initial estimate of the calibration matrix. Learning-based approaches, on the other hand, achieve good calibration performance but suffer from generalization capabilities~\cite{survey} and complex architectures~\cite{LCCNet, SOAC, FusionNet}. In addition, both explicit and learned-based approaches often depend on the presence of features such as rich texture, edges, or lines in the environment, which effectively limits the real-time capabilities of such models. More importantly, none of these methods consider or discuss the detection of miscalibrations. This is an important requirement in real-world applications and for both safety and robustness, miscalibration must be detected before online calibration can be performed.

\subsection{Classification-based Concepts}
Analogous to Section~\ref{sec:regressionbased}, we will categorize the classification-based concepts into methods that require explicit or learned correspondences to detect miscalibrations. However, finding correspondences differs between regression-based and classification-based approaches. The regression-based approaches aim to find correspondences, i.e. matching features to align them and compute the calibration parameters. Classification-based approaches check for spatial similarity of features to discriminate between different modalities. 

\subsubsection*{Explicit Correspondences}
Levinson and Thrun~\cite{Levinson} are the first to specifically consider miscalibration detection before estimating calibration parameters. They use depth discontinuities and image edges to detect misalignment between sensors. This is incorporated into an objective function that indicates whether small adjustments to the current calibration will result in a decrease in the cost function. This objective function measures "edginess" based on LiDAR and image data. 

\subsubsection*{Learned Correspondences}
Recent work in the field of miscalibration detection also tackles this problem by using deep learning techniques. These methods are mainly applied in sensor monitoring techniques, not only in LiDAR-to-camera setups but also in a wider range of systems beyond the field of autonomous driving~\cite{Ma2018, Qian2022}. In the field of autonomous driving, Chen~\etal~\cite{crossmodal} presents a sensor monitoring framework that detects the spatial and temporal misalignment of sensor data as well as single sensor error sources such as image blur, point cloud noise, and perturbation. They train a Siamese network using a contrastive loss to detect sensor and cross-sensor inconsistencies. Wei~\etal~\cite{Wei} introduce a self-checking framework that specifically detects sensor miscalibrations. For feature extraction, they use a patch transformer in the LiDAR and Unet in the camera domain to check for cosine similarity. Their framework implements the LCCNet~\cite{LCCNet} recalibration algorithm. Since they are the first and currently the only ones to perform learning-based miscalibration detection, we use this method as a baseline.  

The role of classification-based concepts in sensor calibration is an important consideration that has not yet received much attention in the literature. It can be considered as a continuous monitoring system and a trigger for the recalibration process or any safety-related modification in the fusion strategy. As introduced in Section~\ref{sec:intro}, for safety reasons, these methods must be designed with resource efficiency and fast inference time in mind. Addressing this research gap, we present a two-stage learning approach and analyze the architecture (see Fig.~\ref{fig:our_architecture} below) with a focus on monitoring for extrinsic calibration errors. We design the network to meet the requirements of model size and inference time.
\section{Method}
\label{sec:method}
\subsection{Fault Injection}\label{sec:faultinjection}
To generate miscalibrated data, we introduce perturbations during the projection of a 3D point cloud into a 2D image plane from the KITTI dataset \cite{KITTI}. First, a 3D point is transformed from the LiDAR coordinate frame to the camera coordinate frame using the transformation matrix $\mathbf{T}_{lid}^{cam}$, which consists of the rotation matrix~$\mathbf{R}^{{cam}}_{{lid}}$ and translation vector~$\mathbf{t}^{{cam}}_{{lid}}$. The transformed point $x$ is then projected onto the rectified, rotated image plane of the i-th camera using the camera projection matrix $\mathbf{P}_{rect}^{(i)}$ and the rectifying rotation matrix of the reference camera $\mathbf{R}_{rect}^{(0)}$. $\mathbf{P}_{rect}^{(i)}$ is the intrinsic calibration matrix with the parameters focal length $f_u, f_y$, the principal point offsets $c_x, c_y$, and the axis skew $\gamma$. The full transformation is shown in Eq. \ref{eq:velo-trafo}.
\begin{equation}
\begin{aligned}
\textbf{y} &= \mathbf{P}_{rect}^{(i)} \cdot \mathbf{R}_{rect}^{(0)} \cdot \mathbf{T}_{lid}^{cam} \cdot \textbf{x} \\
&\hspace{-1em} \\[-1.5em]
&=
\mathbf{P}_{rect}^{(i)}
\cdot
\begin{bmatrix}
\Tilde{\mathbf{R}}^{(0)}_{rect} & 0 \\
0 & 1
\end{bmatrix}
\cdot
\begin{bmatrix}
\mathbf{R}^{{cam}}_{{lid}} & \mathbf{t}^{{cam}}_{{lid}} \\
0 & 1
\end{bmatrix}
\cdot \textbf{x}.
\end{aligned}
\label{eq:velo-trafo}
\end{equation}

To simulate the extrinsic calibration error, we augmented the dataset by introducing perturbations in both the rotational and translational components of this transformation, as shown in Eq. \ref{eq:error-calib}. For the rotational component, we applied errors to the roll $\theta_1$, pitch $\theta_2$, and yaw $\theta_3$ angles by multiplying the transformation matrix by the corresponding rotation matrices $\mathbf{R}_{i, err}$. For the translational component, the error is introduced by adding the translation vector $\mathbf{t}_{err}$ to the relative positions between the LiDAR and camera sensors:
\begin{equation}
    \Tilde{\mathbf{T}}_{lid}^{cam} = \mathbf{T}_{lid}^{cam} \cdot \mathbf{R}_{1,err} \cdot \mathbf{R}_{2,err} \cdot \mathbf{R}_{3,err} + \mathbf{t_{err}}.
    \label{eq:error-calib}
\end{equation}

\subsection{Miscalibration Detection}
Our approach consists of two steps and is shown in Figure~\ref{fig:classifier}. First, we utilize multimodal, pixel-wise contrastive learning to learn the distinct input representations between correctly calibrated and miscalibrated inputs. The embeddings of the frozen encoders are then used for the detection task. This two-step approach using pretext and downstream tasks~\cite{MoCo, misra2020self, chen2021empirical} generally showed promising results compared to end-to-end supervised learning. As described by He \etal \cite{MoCo}, in the pretext task, the model solely learns input feature representations. The downstream task is the primary task of the model's application, in our case, miscalibration detection.

The generation of negative samples presents a significant challenge in contrastive learning. For our problem, however, the constraints imposed by the six degrees of freedom in the relative pose between LiDAR and camera enable the efficient generation of negative samples for contrastive learning. These constraints allow the systematic generation of negative pairs by introducing plausible calibration errors into the dataset, as described in the previous section. Because of its ability to generate discriminative representations for similar and dissimilar data points, contrastive learning is particularly interesting for miscalibration detection.

\subsubsection{Pixel-wise contrastive learning}\label{sec:contrastivelearning}
For those reasons, we chose the model architecture based on contrastive learning with two-stream encoders, similar to the architecture proposed by Hadsell \etal\cite{hadsell2006dimensionality}. However, in contrast to the more common contrastive learning models that utilize augmentations of RGB images, such as \cite{hadsell2006dimensionality, SimCLR, MoCo}, our model processes both RGB images and LiDAR data, which could be seen as different views of the same scene or pair, analogous to how contrastive learning typically uses multiple views of an image. 

As with other contrastive methods, the model must learn to distinguish between positive and negative pairs. In this framework, positive (correct) pairs consist of RGB images and corresponding LiDAR data. Negative (incorrect) pairs, on the other hand, are the data samples with misalignments between images and LiDAR points due to miscalibrations.

The overall pipeline of the model is shown in Figure~\ref{fig:classifier}. First, the 3D LiDAR points are projected into the 2D image plane. In this step, miscalibration errors are introduced through the miscalibrated transformation matrix as explained in Section~\ref{sec:faultinjection}. Each modality is processed by separate encoders: the RGB images are processed by an image encoder, and the LiDAR data by a LiDAR encoder. These encoders share a common architecture, based on the ResNet18 model, but differ in the input channels: three channels for RGB images and one channel for LiDAR data, representing depth information. Due to different modalities and input channels, unlike unimodal contrastive learning, weight sharing of the encoders cannot be applied.

To learn a feature representation, we only use the first two residual blocks of the ResNet18 architecture for the encoders. This compact model reduces computational load, making it suitable for real-time applications, such as autonomous driving, where resources are often constrained. Furthermore, smaller models may preserve more spatial details, which can be advantageous when comparing the embeddings \cite{crossmodal}, which will be discussed in more detail in Section~\ref{sec:discussion}. Additionally, given the inherent sparsity of LiDAR data, using a compact model ensures that meaningful features are extracted even when depth information is limited. While Wang~\etal~\cite{crossmodal} use a sparsity mask to account for the inherent sparsity of LiDAR data, our framework achieves strong results without the additional step indicating masking has negligible effect.

After feature extraction, the resulting embeddings from the image and LiDAR data samples are compared using a contrastive loss function to measure the distance of the embeddings. The key difference in our approach compared to common contrastive learning methods is the employment of a pixel-wise comparison, which ensures that even calibration errors at the pixel level are considered. The pixel-wise contrastive loss function is adapted from the standard contrastive loss function \cite{hadsell2006dimensionality}, but instead of projecting the image and LiDAR embeddings into a one-dimensional vector for comparison, the spatial resolution of the embeddings is maintained. The contrastive loss function is defined as follows:

\begin{equation}
\begin{split}
L = \frac{1}{NHW} \sum_{n,h,w} \big[ & (1-y) \, D_{nhw}^2 \\
& + y \cdot \max(0, m - D_{nhw})^2 \big],
\end{split}
\label{eq:pixel-CL-loss}
\end{equation}
where $N$ is the batch size, $H$ and $W$ are the dimension of the embeddings, and $m$ is the margin. The pixel-wise distance is defined as $D_{nhw} = \sqrt{\sum_{c=0}^C (E_{I,nchw} - E_{L,nchw})^2}$. The label y = 1 for the incorrect pairs (miscalibrated inputs) and y = 0 for the correct pairs (calibrated inputs).

Unlike previous work \cite{Wei}, which uses a cost volume layer to compare and constrain the similarity between the LiDAR and RGB features in the embedding layer before the decoder, we directly optimize both similarity and dissimilarity as the training objective. Additionally, our approach efficiently learns (dis)similarity and alignment of the input representations by omitting the decoder.

\subsubsection{Detection task} 
After completing the pretext task of training the encoders using our multimodal contrastive learning framework, the learned representations are applied to the subsequent miscalibration detection task. This task determines whether the given sensor inputs are correctly calibrated or miscalibrated based on the embeddings learned from the encoders and is therefore designed as a binary classification task.

\begin{figure*}
    \centering
    \includegraphics[width=\textwidth]{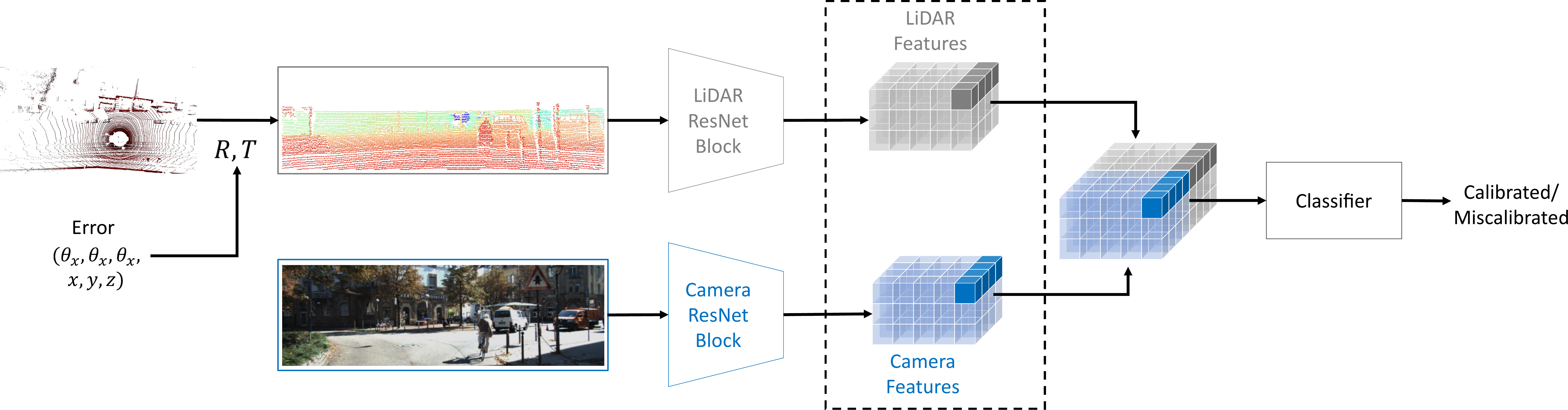}
    \caption{Pipeline for the classification task within the sensor miscalibration detection. Both image and LiDAR inputs are processed through ResNet blocks, with feature extraction performed for each modality. The ResNet blocks are frozen, and the extracted features are then concatenated and passed to a classifier, which predicts whether miscalibration is detected or not. The feature embeddings inside the dotted box are trained in the first stage using contrastive loss.}
    \label{fig:classifier}
    \vspace{-12pt}
\end{figure*}

As illustrated in Fig.~\ref{fig:classifier}, the embeddings from both the LiDAR and image encoders are concatenated along the channel dimension, forming a spatially aligned feature representation that integrates information from both sensor modalities. This concatenation captures the relationships between the LiDAR and corresponding image embeddings, which is crucial for accurately detecting miscalibrations. 

The classifier head consists of three 3x3 convolutional layers to further reduce the dimension, followed by global average pooling and three fully connected (MLP) layers with 512 and two 216 neurons. The final layer uses a sigmoid activation function to produce a probability score that indicates whether the sensor inputs are miscalibrated (class 1) or calibrated (class 0). Binary cross-entropy is used as the loss function.

\section{Implementation Details}
\label{sec:experiments}
In this section, we outline the implementation details of our approach, starting with the dataset perturbations used to simulate various calibration errors. We then describe the training configuration for our architecture.
\subsection{Datasets}
To train and evaluate our model, we used the KITTI Odometry dataset, which is also used by Wei~\etal~\cite{Wei}. For comparability, we used the same sequences as in the previous works -- the sequences from '01' to '20' comprising 39,011 frames for training and '00' with 4541 frames for the test set. Table~\ref{tab:error_ranges} shows the range of errors included in the calibrated and miscalibrated datasets. The calibrated dataset includes small allowed shifts as tolerance to simulate environmental noise and relative deviations between different calibrations in the KITTI dataset. The tolerances are based on benchmarks that investigate the robustness of miscalibration on object detection performance~\cite{Yu2023, Fuerst2024}. Accordingly, the ranges are chosen with the expectation that our model will detect even more challenging misalignments. The miscalibrated dataset contains larger perturbations in both translational and rotational components, which reflect the errors we want to detect. In addition, we also introduced a small margin between the calibrated and miscalibrated datasets. Without this margin, the miscalibrated and calibrated data points at the interval boundaries would be similar or even identical. This similarity leads the model to encounter nearly identical or the same input labeled as miscalibrated and calibrated, which introduces ambiguity and ultimately degrades data quality.
\begin{table}[h!]
  \centering
  \begin{tabular}{@{}lcc@{}}
    \toprule
    \textbf{Datasets} & \parbox{2cm}{\centering \textbf{Translation Error} \\ {\textbf{[m]}}} & \parbox{2cm}{\centering \textbf{Rotation Error} \\ {\textbf{[$^\circ$]}}} \\
    \midrule
    \textbf{Calibrated} & \([0, 0.02]\) & \([0, 0.3]\) \\
    \textbf{Miscalibrated} & \([0.04, 0.1]\) & \([0.5, 5]\) \\
    \bottomrule
  \end{tabular}
  \caption{Absolute error ranges of calibrated and miscalibrated dataset for training and validation.}
  \label{tab:error_ranges}
\end{table}
During the evaluation, we first introduce small noise to the calibrated dataset to monitor whether negligible errors will trigger false alarms of miscalibration. We evaluate the model for various miscalibration errors for the test as described in Table \ref{tab:test_datasets}. The goal of using these datasets is to evaluate the model’s performance in detecting different error modes with different magnitudes and the generalization ability to detect unseen errors. Lastly, we exclude small intervals between the calibrated and miscalibrated dataset, assuming that the model is allowed to overlook errors within the range.

\begin{table}[h!]
  \centering
  \begin{tabular}{@{}lcc@{}}
    \toprule
    \textbf{Datasets} & \parbox{2cm}{\centering \textbf{Translation Error} \\ {\textbf{[m]}}} & \parbox{2cm}{\centering \textbf{Rotation Error} \\ {\textbf{[$^\circ$]}}} \\
    \midrule
    \textbf{Noise} & \([0, 0.005]\) & \([0, 0.1]\) \\
    \textbf{Miscalibrated} & \([0.04, 0.1]\) & \([0.5, 5]\) \\
    \textbf{Unseen} & \([0.1, 0.2]\) & \([5, 10]\) \\
    \textbf{All Errors} & \([0.1, 0.2]\) & \([0.5, 1]\) \\
    \textbf{Rot hard} & 0 & \([0.5, 1]\) \\
    \textbf{Rot easy} & 0 & \([1, 5]\) \\
    \textbf{Trans hard} & \([0.04, 0.1]\) & 0 \\
    \textbf{Trans easy} & \([0.1, 0.2]\) & 0 \\
    \bottomrule
  \end{tabular}
  \caption{Absolute error ranges for different dataset configurations for the evaluation and they cover different error modes and magnitudes.}
  \label{tab:test_datasets}
\vspace{-10pt}
\end{table}

\subsection{Training Configurations}
Each phase of training was conducted on an A100 GPU with a batch size of 64, evenly split into 32 calibrated and 32 miscalibrated samples to maintain class balance. The AdamW optimizer was used, with an initial learning rate of 0.001 and a weight decay of 0.05. The learning rate was reduced to 1e-4 after 30 epochs for fine-tuning, and training continued for a total of 50 epochs. For contrastive loss, a margin $m=4$ was selected.

\section{Results and Discussion}
\label{sec:results}
This chapter evaluates the model’s performance across various miscalibration conditions, real-time capability, and the learned representations. For the model performance, we use the evaluation metrics accuracy, precision, and recall to measure the prediction outputs. In terms of miscalibration detection, a high precision score reduces the rate of false positives, which would trigger false alerts for system recalibrations. Higher recall values ensure that most miscalibrations are detected, minimizing the risk of missing actual calibration errors. This is especially important for safety-critical systems like autonomous driving, where undetected miscalibrations could degrade sensor performance and compromise safety. Furthermore, we utilized Centered Kernel Alignment analysis to evaluate the resulting embeddings across layers. CKA analysis is a technique used to compare the similarity of embeddings (i.e., representations) from different layers of the same or different neural networks \cite{cka-kornblith2019similarity, cka-cortes2012algorithms, raghu2021vision}. This analysis provides insights into how the model processes and represents input data at various stages, allowing us to better understand the internal representation learned by each layer.

\subsection{Miscalibration Detection Performance}
The results on the miscalibration performance are shown in Table~\ref{tab:performance_metrics_compare} and~\ref{tab:eval_all}. We first compare the performance of our model to LCCNet~\cite{LCCNet} and Wei~\etal~\cite{Wei}. 

\begin{table}[h!]
  \centering
  \begin{tabular}{@{}llccc@{}}
    \toprule
    \textbf{Metrics} & \parbox{1.2cm}{\centering \textbf{Methods}} & \parbox{1.5cm}{\centering \textbf{All Errors}}  & \parbox{1cm}{\centering \textbf{Rot Hard}} & \parbox{1cm}{\centering \textbf{Trans Easy}} \\
    \midrule
    \textbf{Accuracy} & LCCNet & 90.91\% & 86.48\% & 90.44\% \\
                      & Wei & 95.13\% & 86.28\% & 92.05\% \\
                      & Ours & \textbf{99.08}\% & \textbf{99.00}\% & \textbf{99.97}\% \\
    \midrule
    \textbf{Precision} & LCCNet & 88.79\% & 78.69\% & 85.63\% \\
                       & Wei & 92.02\% & 78.24\% & 86.59\% \\
                       & Ours & \textbf{100.00}\% & \textbf{100.00}\% & \textbf{100.00}\% \\
    \midrule
    \textbf{Recall} & LCCNet & 94.04\% & 99.51\% & 97.38\% \\
                    & Wei & \textbf{99.05}\% & \textbf{99.96}\% & 99.65\% \\
                    & Ours & 98.17\% & 97.99\% & \textbf{99.93}\% \\
    \bottomrule
  \end{tabular}
  \caption{Results on KITTI Odometry and comparison with existing methods.}
  \label{tab:performance_metrics_compare}
  \vspace{-8pt}
\end{table}

Compared to the existing methods, our model shows better accuracy and precision performance with an accuracy of around 99\% and a precision of 100\% highlighting the robustness of our approach. A robust miscalibration detection refers to a system that does not trigger false positive recalibration for example. Considering that current online calibration algorithms do not achieve target-based calibration accuracies, avoiding false positive recalibrations also contributes to the overall safety of the system. In terms of recall, our method shows a slightly lower performance in the range of around 1\% for combined and 2\% for rotational errors. The translational errors, however, are more accurately detected by our approach. Looking at the overall performance presented in Table~\ref{tab:performance_metrics_compare}, the existing methods show over-sensitivity to miscalibration errors that caused reduced precision and poor trade-off between the precision and recall metrics. Note that LCCNet is not specifically provided with a miscalibration detection, but was extended by~\cite{Wei} with a classification module to be able to benchmark against existing methods. Table~\ref{tab:inference_time_model_size} shows that our model has significantly smaller model size and inference time. This highlights the real-time capability of our model, while achieving good performance in the evaluation metrics.

\begin{table}[h!]
  \centering
  \begin{tabular}{@{}lcc@{}}
    \toprule
    \textbf{Methods} & \parbox{3cm}{\centering \textbf{Inference Time} \\ {\textbf{[ms]}}} & \parbox{2cm}{\centering \textbf{Model Size} \\ {\textbf{[M]}}} \\
    \midrule
    LCCNet & 97.1 & 210 \\
    Wei & 160.7 & 49 \\
    Ours & \textbf{26.5} & \textbf{28} \\
    \bottomrule
  \end{tabular}
  \caption{Comparison of the inference time and model size evaluated on an Nvidia A100 GPU. The inference time results were obtained according to the evaluation method presented in~\cite{Wei}. Model size refers to the number of parameters of the models in millions.}
  \label{tab:inference_time_model_size}
\end{table}
Table~\ref{tab:eval_all} extends the error modes and magnitudes (referring to the results of the first three rows). Compared to Table~\ref{tab:performance_metrics_compare}, additional modes such as unseen errors as well as larger rotational errors and more challenging translational errors are evaluated. We observe an increase in performance for larger rotational calibration errors and maintain high detection performance for various sets of miscalibrated datasets including unseen calibration errors. This highlights the strength of our framework to capture very challenging translational errors and to distinguish between rotational and translational errors for the combination of errors.
\begin{table*}[h!]
  \centering
  \begin{tabular}{@{}llccccccc@{}}
    \toprule
    \textbf{Metrics} & \textbf{Encoders} & \textbf{Miscalibrated} & \textbf{Unseen} & \textbf{Rot Easy} & \textbf{Rot Hard} & \textbf{Trans Easy} & \textbf{Trans Hard} \\
    \midrule
    \textbf{Accuracy} &   & \textbf{99.42\%} & \textbf{99.97\%} & \textbf{99.99\%} & \textbf{99.00\%} & \textbf{99.97\%} & \textbf{99.22\%} \\
    \textbf{Precision} & ResNet18-Small & \textbf{100.00\%} & \textbf{100.00\%} & \textbf{100.00\%} & \textbf{100.00\%} & \textbf{100.00\%} & \textbf{100.00\%} \\
    \textbf{Recall} &   & \textbf{98.84\%} & \textbf{99.93\%} & \textbf{99.98\%} & \textbf{97.99\%} & \textbf{99.93\%} & 98.44\% \\
    \midrule
    \textbf{Accuracy} &  & 98.21\% & 99.36\% & 99.11\% & 94.02\% & 99.33\% & 98.56\% \\
    \textbf{Precision} & ResNet18-All & 99.36\% & 99.37\% & 99.37\% & 99.30\% & 99.37\% & 97.75\% \\
    \textbf{Recall} &   & 97.05\% & 99.35\% & 98.84\% & 88.66\% & 99.29\% & \textbf{99.36}\% \\
    \bottomrule
  \end{tabular}
    \caption{Performance metrics of different encoders for different test conditions, as shown in Table~\ref{tab:test_datasets}.}
  \label{tab:eval_all}
  \vspace{-10pt}
\end{table*}

\subsection{CKA Analysis}
\begin{figure}[!h]
    \centering
    \begin{subfigure}{0.49\linewidth}
        \centering
        \includegraphics[width=\linewidth]{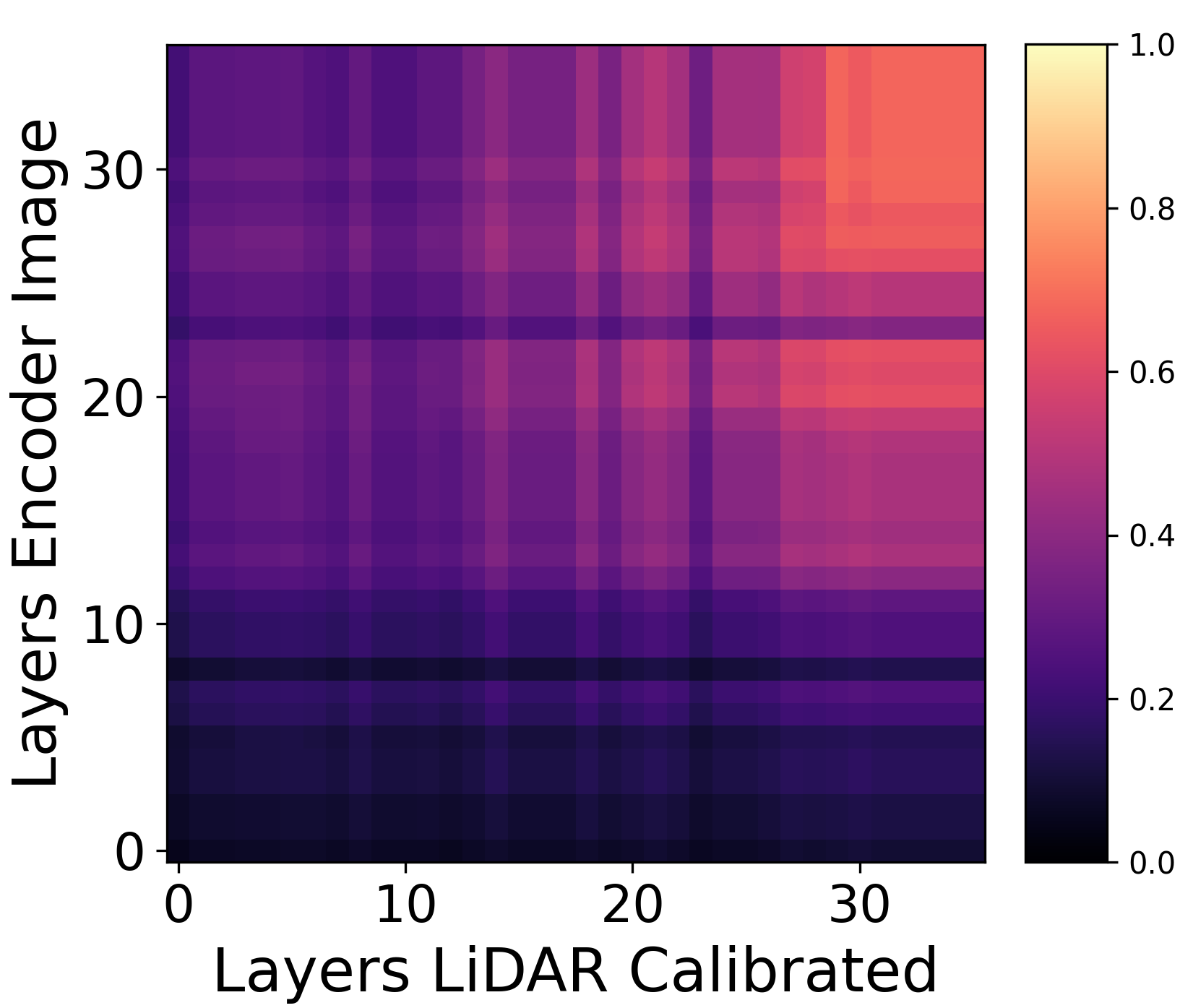}
        \caption{}
        \label{fig:calibrated}
    \end{subfigure}
    \hfill
    \begin{subfigure}{0.49\linewidth}
        \centering
        \includegraphics[width=\linewidth]{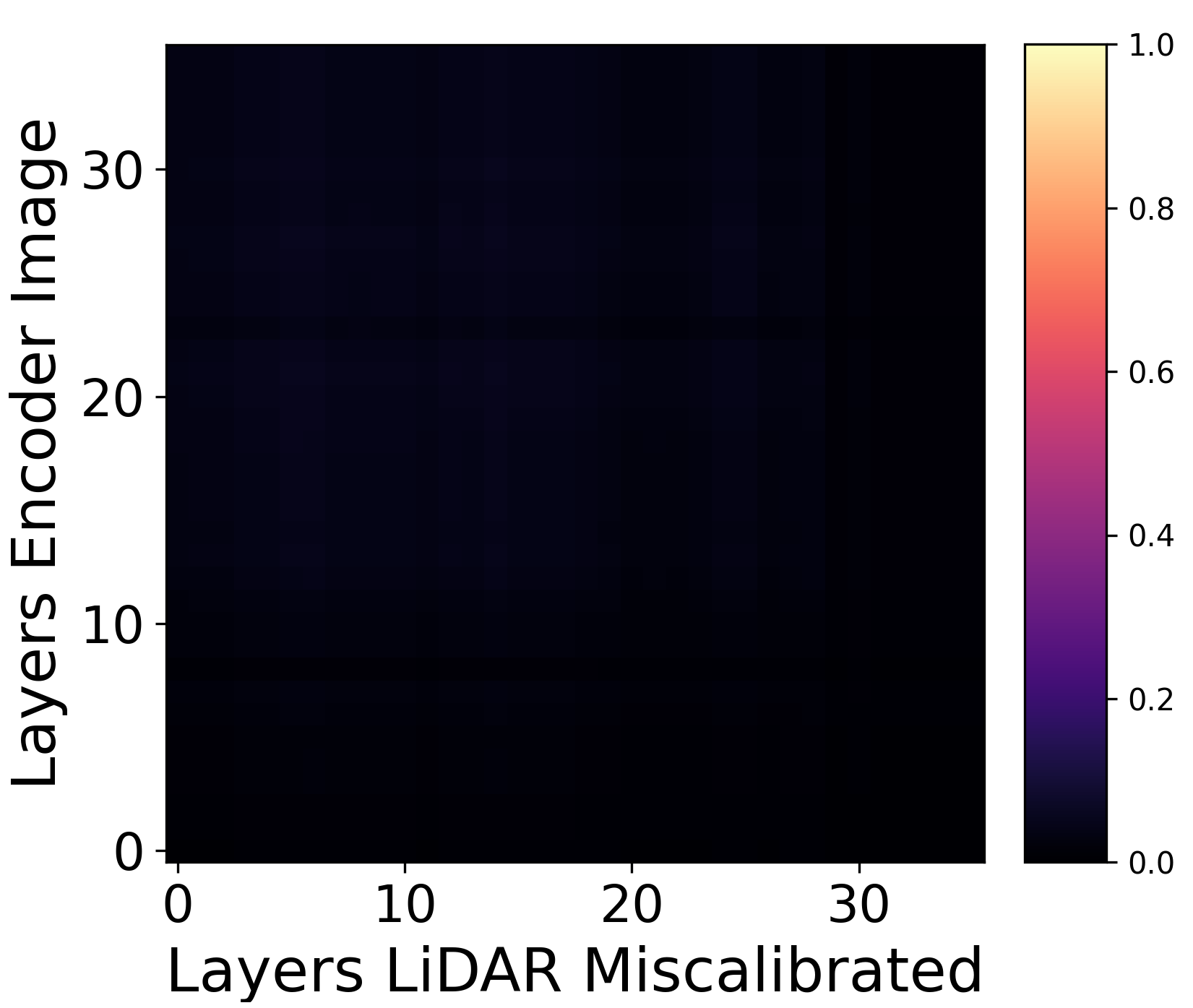}
        \caption{}
        \label{fig:miscalibrated}
    \end{subfigure}
    \caption{CKA analysis between LiDAR and image encoders, evaluated on calibrated~\ref{fig:calibrated} and miscalibrated input~\ref{fig:miscalibrated}. A CKA value of 1 indicates identical feature embeddings between the encoders in a given layer, whereas a value close to 0 means minimal similarity in their representations.}
    \label{fig:cka}
    \vspace{-12pt}
\end{figure}

We utilized CKA analysis to evaluate the similarity of feature representations learned by the image and LiDAR encoder in different layers. As depicted in Figure \ref{fig:calibrated}, the CKA values across the layers of the two encoders when taking the calibrated dataset as inputs shows that the deeper layers of the network exhibit a higher degree of similarity. When analyzing miscalibrated samples (Figure \ref{fig:miscalibrated}), the CKA values remain generally low across most layers, indicating a clear difference between feature embeddings generated from miscalibrated data. 
This shows that the model can effectively differentiate between calibrated and miscalibrated inputs, as miscalibration distorts the feature alignment between the two encoders. This behavior is particularly crucial for detecting sensor misalignment.

\subsection{Intrinsic Miscalibration}
We evaluate the performance of our algorithm in the presence of intrinsic calibration errors in $\mathbf{P}_{rect}^{(i)}$. Specifically, we introduce relative (percentage) errors in the focal lengths ($f_u, f_v$) and principal point coordinates ($c_u, c_v$) by $n\%$. For the skew $\gamma$ we introduce an error of $n\%$ of $f_u$, where $n=10-20$, $n=5-10$, $n=3-5$ are denoted as easy, medium and hard, respectively. The results are shown in Table~\ref{tab:intrinsics}. The performance of our approach decreases only slightly for the accuracy and precision metrics when intrinsic calibration errors are introduced. However, the results of the recall metric indicate that not all intrinsic calibration errors can be detected. Nevertheless, we still achieve significantly high scores for all metrics, considering that the framework is trained only on extrinsic errors and therefore showing good generalization capability for unseen error patterns.

\begin{table}[h!]
  \centering
  \begin{tabular}{@{}lccc@{}}
    \toprule
\textbf{Metrics} & \textbf{easy}  & \textbf{medium} & \textbf{hard} \\
    \midrule
    \textbf{Accuracy} & 96\% & 94\% & 94\% \\
    \textbf{Precision}  & 99\% & 99\% & 99\% \\
    \textbf{Recall} & 93\% & 88\% & 88\% \\
    \bottomrule
  \end{tabular}
  \caption{Results of the miscalibration detection for intrinsic calibration errors.}
  \label{tab:intrinsics}
  \vspace{-8pt}
\end{table}

We now want to discuss the practical implications of this work. It is clear that relying solely on miscalibration detection is not sufficient for autonomous driving tasks. The main practical implication lies in the continuous monitoring of the calibration state of the system, which can be achieved with the computational resources and inference time provided by this method. It should be noted, however, that after recalibrating sensors using regression-based methods, zero calibration errors cannot be guaranteed. Especially, when intrinsic errors are compensated using extrinsic calibration methods. This becomes more severe with axis skew errors, which cannot be compensated using extrinsic recalibration methods. In such cases, our approach can quickly trigger recalibration when initial miscalibration occurs due to real-world effects such as sensor drift or temperature changes, or when corrections are needed after an erroneous recalibration.
\section{Ablation Study}
\label{sec:discussion}
This chapter explores the ablation of encoders by using the entire ResNet18 model (in the following denoted as ResNet18-All). For this analysis we exclude its average pooling and fully connected layers to account for the deeper encoders, which result in reduced feature dimensions. We also modified the classifier head by simplifying it to a single 3x3 convolutional layer.

\subsection{CKA Analysis}
\begin{figure}[!h]
    \centering
    \begin{subfigure}{0.49\linewidth}
        \centering
        \includegraphics[width=\linewidth]{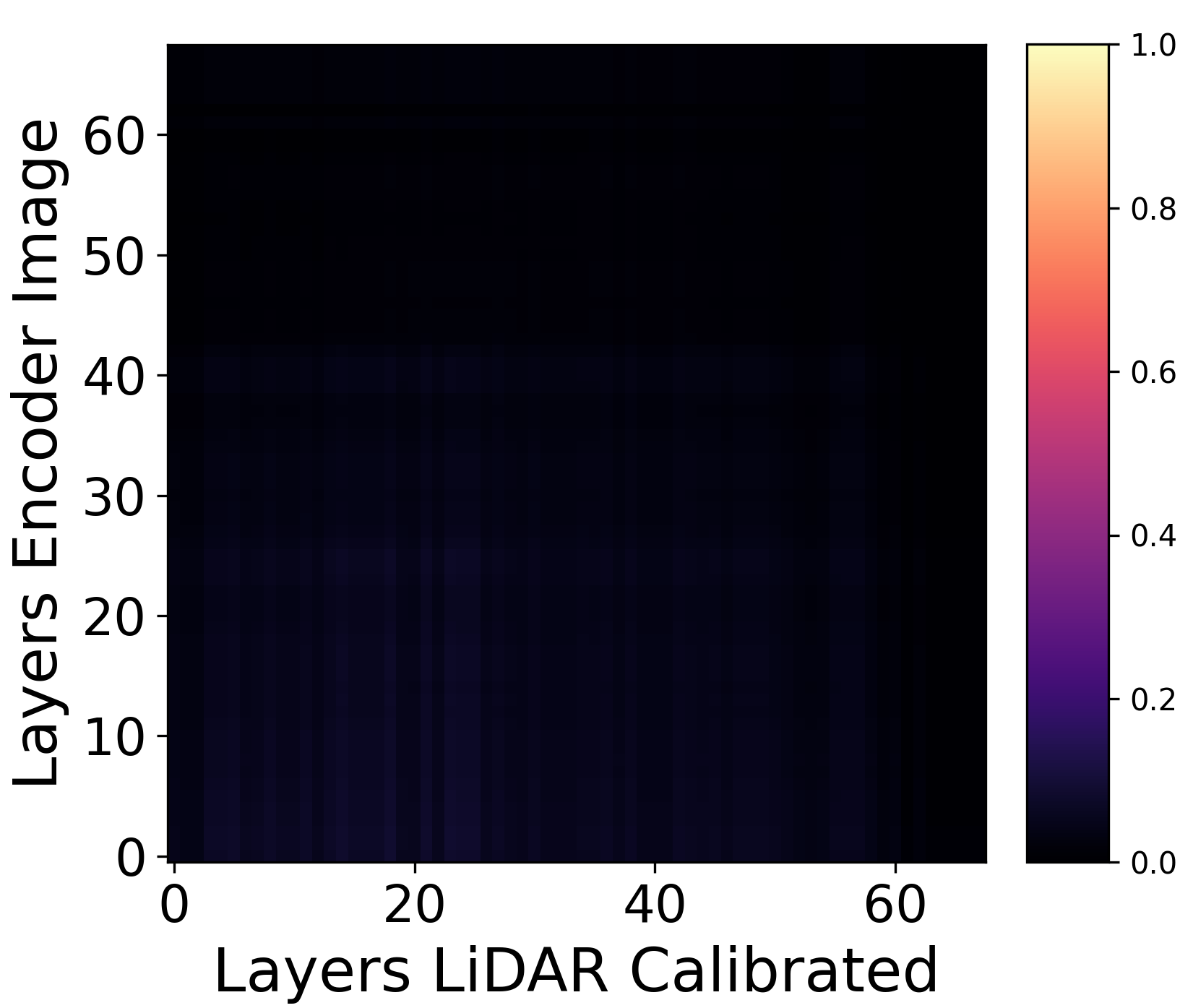}
        \caption{}
        \label{fig:calibrated_all}
    \end{subfigure}
    \hfill
    \begin{subfigure}{0.49\linewidth}
        \centering
        \includegraphics[width=\linewidth]{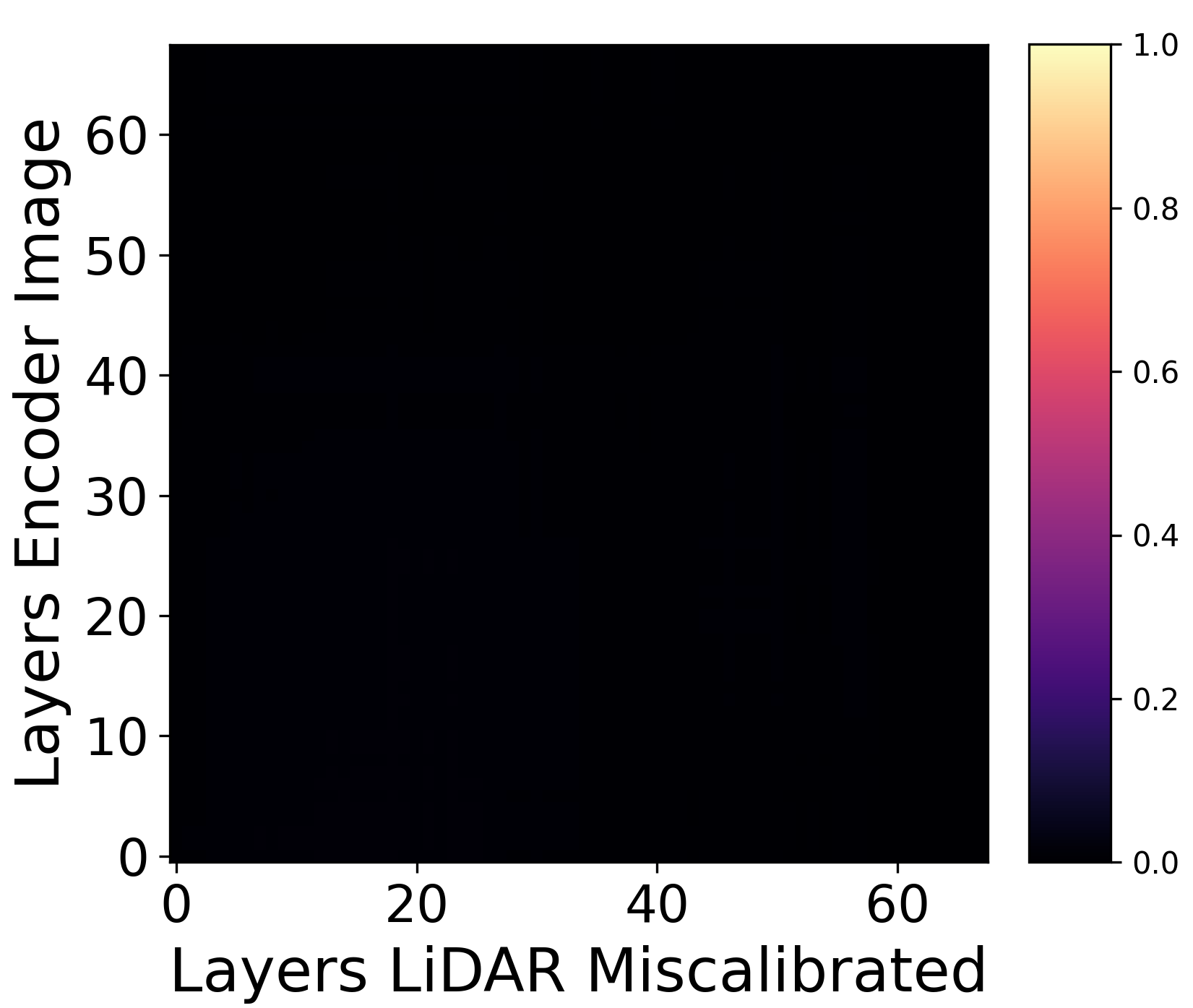}
        \caption{}
        \label{fig:miscalibrated_all}
    \end{subfigure}
    \caption{CKA analysis between LiDAR and image encoders of whole ResNet18 architecture evaluated on calibrated~\ref{fig:calibrated_all} and miscalibrated input~\ref{fig:miscalibrated_all}.}
    \label{fig:cka_all}
    \vspace{-12pt}
\end{figure}

The results in Fig.~\ref{fig:cka_all} show that the learned embeddings between the image and the LiDAR encoder trained with ResNet18-All have very low similarity. The reasons are twofold: In the loss function, we only imposed the distance constraint and not the similarity. 
It is important to note that a small distance between embeddings, as measured by the contrastive loss, does not necessarily imply that the embeddings are similar in terms of their feature representations. In addition, since we could not apply weight sharing, the features of two encoders across different layers are not aligned and most likely learn modality-specific features.

\subsection{Miscalibration Detection Performance}
Table~\ref{tab:eval_all} describes the influence of using ResNet18-All as encoders on the performance of the miscalibration detection. In contrast to ResNet18-Small, a lower overall performance in almost all calibration conditions and metrics can be observed. Both the CKA analysis as well as the evaluation metrics validates that our smaller model indeed preserves relevant spatial details. The results indicate that the smaller ResNet18-Small encoder not only achieves better accuracy, precition, and recall but also allows us to employ a more compact model architecture.

\section{Conclusion}
\label{sec:conclusion}
We introduce a novel framework for safe and real-time miscalibration detection in LiDAR-camera sensor setups, focusing on binary classification to distinguish calibrated from miscalibrated states. Our approach leverages contrastive learning and a simple and efficient architecture to achieve state-of-the-art performance. Through our comprehensive analysis, we identify the depth at which latent features should be compared, enabling us to simplify the feature extraction and matching layers without compromising accuracy. Our experiments demonstrate that our method not only achieves state-of-the-art performance in various miscalibration scenarios, but also requires less computational resources and inference time to detect miscalibration, making it well suited for real-time applications. 


\bibliography{MiscalibrationDetection.bib}

\begin{thebibliography}{10}
\providecommand{\url}[1]{#1}
\csname url@samestyle\endcsname
\providecommand{\newblock}{\relax}
\providecommand{\bibinfo}[2]{#2}
\providecommand{\BIBentrySTDinterwordspacing}{\spaceskip=0pt\relax}
\providecommand{\BIBentryALTinterwordstretchfactor}{4}
\providecommand{\BIBentryALTinterwordspacing}{\spaceskip=\fontdimen2\font plus
\BIBentryALTinterwordstretchfactor\fontdimen3\font minus \fontdimen4\font\relax}
\providecommand{\BIBforeignlanguage}[2]{{%
\expandafter\ifx\csname l@#1\endcsname\relax
\typeout{** WARNING: IEEEtran.bst: No hyphenation pattern has been}%
\typeout{** loaded for the language `#1'. Using the pattern for}%
\typeout{** the default language instead.}%
\else
\language=\csname l@#1\endcsname
\fi
#2}}
\providecommand{\BIBdecl}{\relax}
\BIBdecl

\bibitem{Wang2023}
L.~Wang \emph{et~al.}, ``Multi-modal 3d object detection in autonomous driving: A survey and taxonomy,'' \emph{IEEE Transactions on Intelligent Vehicles}, vol.~8, pp. 3781--3798, 7 2023.

\bibitem{Yu2023}
K.~Yu \emph{et~al.}, ``Benchmarking the robustness of lidar-camera fusion for 3d object detection,'' in \emph{2023 IEEE/CVF Conference on Computer Vision and Pattern Recognition Workshops (CVPRW)}, 2023.

\bibitem{Dong2023}
\BIBentryALTinterwordspacing
Y.~Dong \emph{et~al.}, ``Benchmarking robustness of 3d object detection to common corruptions in autonomous driving,'' in \emph{2023 IEEE/CVF Conference on Computer Vision and Pattern Recognition (CVPR)}, vol.~8, 3 2023. [Online]. Available: \url{http://arxiv.org/abs/2303.11040}
\BIBentrySTDinterwordspacing

\bibitem{Bai2022}
X.~Bai \emph{et~al.}, ``Transfusion: Robust lidar-camera fusion for 3d object detection with transformers,'' in \emph{2022 IEEE/CVF Conference on Computer Vision and Pattern Recognition (CVPR)}, 2022.

\bibitem{Xie2023}
\BIBentryALTinterwordspacing
Y.~Xie \emph{et~al.}, ``Sparsefusion: Fusing multi-modal sparse representations for multi-sensor 3d object detection,'' in \emph{ICCV}, 2023. [Online]. Available: \url{https://github.com/yichen928/SparseFusion.}
\BIBentrySTDinterwordspacing

\bibitem{Song2023}
Z.~Song \emph{et~al.}, ``Graphalign: Enhancing accurate feature alignment by graph matching for multi-modal 3d object detection,'' in \emph{ICCV}, 2023.

\bibitem{Liu2023}
\BIBentryALTinterwordspacing
Z.~Liu \emph{et~al.}, ``Bevfusion: Multi-task multi-sensor fusion with unified bird's-eye view representation,'' in \emph{IEEE International Conference on Robotics and Automation}, 5 2023. [Online]. Available: \url{http://arxiv.org/abs/2205.13542}
\BIBentrySTDinterwordspacing

\bibitem{survey}
P.~An \emph{et~al.}, ``Survey of extrinsic calibration on lidar-camera system for intelligent vehicle: Challenges, approaches, and trends,'' \emph{IEEE Transactions on Intelligent Transportation Systems}, pp. 1--25, 7 2024.

\bibitem{LCCNet}
\BIBentryALTinterwordspacing
X.~Lv \emph{et~al.}, ``Lccnet: Lidar and camera self-calibration using cost volume network,'' in \emph{2021 IEEE/CVF Conference on Computer Vision and Pattern Recognition Workshops (CVPRW)}, 2021. [Online]. Available: \url{https://github.com/LvXudong-HIT/LCCNet}
\BIBentrySTDinterwordspacing

\bibitem{CalibDepth}
J.~Zhu, J.~Xue, and P.~Zhang, ``Calibdepth: Unifying depth map representation for iterative lidar-camera online calibration,'' in \emph{IEEE International Conference on Robotics and Automation}, vol. 2023-May.\hskip 1em plus 0.5em minus 0.4em\relax Institute of Electrical and Electronics Engineers Inc., 2023, pp. 726--733.

\bibitem{Yuan2021}
C.~Yuan \emph{et~al.}, ``Pixel-level extrinsic self calibration of high resolution lidar and camera in targetless environments,'' \emph{IEEE Robotics and Automation Letters}, vol.~6, pp. 7517--7524, 10 2021.

\bibitem{PBACalib}
F.~Chen \emph{et~al.}, ``Pbacalib: Targetless extrinsic calibration for high-resolution lidar-camera system based on plane-constrained bundle adjustment,'' \emph{IEEE Robotics and Automation Letters}, vol.~8, pp. 304--311, 1 2023.

\bibitem{CRLF}
\BIBentryALTinterwordspacing
T.~Ma, Z.~Liu, G.~Yan, and Y.~Li, ``Crlf: Automatic calibration and refinement based on line feature for lidar and camera in road scenes,'' 3 2021. [Online]. Available: \url{http://arxiv.org/abs/2103.04558}
\BIBentrySTDinterwordspacing

\bibitem{Ishikawa2024}
R.~Ishikawa \emph{et~al.}, ``Lidar-camera calibration using intensity variance cost,'' in \emph{IEEE International Conference on Robotics and Automation (ICRA)}.\hskip 1em plus 0.5em minus 0.4em\relax Institute of Electrical and Electronics Engineers Inc., 2024, pp. 10\,688--10\,694.

\bibitem{SemAlign}
Z.~Liu \emph{et~al.}, ``Semalign: Annotation-free camera-lidar calibration with semantic alignment loss,'' in \emph{IEEE International Conference on Intelligent Robots and Systems}.\hskip 1em plus 0.5em minus 0.4em\relax Institute of Electrical and Electronics Engineers Inc., 2021, pp. 8845--8851.

\bibitem{Luo2024}
Z.~Luo \emph{et~al.}, ``Zero-training lidar-camera extrinsic calibration method using segment anything model,'' in \emph{IEEE International Conference on Robotics and Automation}.\hskip 1em plus 0.5em minus 0.4em\relax Institute of Electrical and Electronics Engineers Inc., 2024, pp. 14\,472--14\,478.

\bibitem{SGCalib}
Z.~Lin \emph{et~al.}, ``Sgcalib: A two-stage camera-lidar calibration method using semantic information and geometric features,'' in \emph{IEEE International Conference on Robotics and Automation (ICRA)}.\hskip 1em plus 0.5em minus 0.4em\relax Institute of Electrical and Electronics Engineers Inc., 2024, pp. 14\,527--14\,533.

\bibitem{Schneider2017}
N.~Schneider \emph{et~al.}, ``Regnet: Multimodal sensor registration using deep neural networks,'' in \emph{2017 IEEE Intelligent Vehicles Symposium (IV)}.\hskip 1em plus 0.5em minus 0.4em\relax IEEE, 2017, p. 1927.

\bibitem{CalibNet}
I.~Ganesh, R.~K. Ram, K.~Murthy, and K.~M. Kirshna, ``Calibnet: Geometrically supervised extrinsic calibration using 3d spatial transformer networks,'' in \emph{2018 IEEE/RSJ International Conference on Intelligent Robots and Systems (IROS)}.\hskip 1em plus 0.5em minus 0.4em\relax IEEE, 2018.

\bibitem{RGGNet}
K.~Yuan, Z.~Guo, and Z.~J. Wang, ``Rggnet: Tolerance aware lidar-camera online calibration with geometric deep learning and generative model,'' \emph{IEEE Robotics and Automation Letters}, vol.~5, pp. 6956--6963, 10 2020.

\bibitem{FusionNet}
G.~Wang, J.~Qiu, and Y.~Guo, ``Fusionnet: Coarse-to-fine extrinsic calibration network of lidar and camera with hierarchical point-pixel fusion,'' in \emph{IEEE International Conference on Robotics and Automation (ICRA)}, 2022.

\bibitem{Hu2024}
X.~Hu \emph{et~al.}, ``Lidar-camera extrinsic calibration with hierachical and iterative feature matching,'' in \emph{IEEE International Conference on Robotics and Automation (ICRA)}.\hskip 1em plus 0.5em minus 0.4em\relax Institute of Electrical and Electronics Engineers Inc., 2024, pp. 16\,691--16\,697.

\bibitem{SOAC}
Q.~Herau, N.~Piasco, M.~Bennehar, L.~Roldão, D.~Tsishkou, C.~Migniot, P.~Vasseur, and C.~Demonceaux, ``Soac: Spatio-temporal overlap-aware multi-sensor calibration using neural radiance fields,'' in \emph{2024 IEEE/CVF Conference on Computer Vision and Pattern Recognition (CVPR)}.\hskip 1em plus 0.5em minus 0.4em\relax Institute of Electrical and Electronics Engineers Inc., 2024.

\bibitem{Levinson}
J.~Levinson and S.~Thrun, ``Automatic online calibration of cameras and lasers,'' in \emph{Robotics: Science and Systems}, 2013.

\bibitem{Ma2018}
M.~Ma \emph{et~al.}, ``Deep coupling autoencoder for fault diagnosis with multimodal sensory data,'' \emph{IEEE Transactions on Industrial Informatics}, vol.~14, pp. 1137--1145, 3 2018.

\bibitem{Qian2022}
J.~Qian \emph{et~al.}, ``A review on autoencoder based representation learning for fault detection and diagnosis in industrial processes,'' \emph{Chemometrics and Intelligent Laboratory Systems}, vol. 231, 2022.

\bibitem{crossmodal}
Y.~Chen \emph{et~al.}, ``Cross-modal matching cnn for autonomous driving sensor data monitoring,'' in \emph{Proceedings of the IEEE International Conference on Computer Vision Workshop}, vol. 2021-October.\hskip 1em plus 0.5em minus 0.4em\relax Institute of Electrical and Electronics Engineers Inc., 2021, pp. 3103--3112.

\bibitem{Wei}
P.~Wei \emph{et~al.}, ``Online lidar-camera extrinsic parameters self-checking and recalibration,'' \emph{Measurement Science and Technology}, vol.~35, 10 2024.

\bibitem{KITTI}
\BIBentryALTinterwordspacing
A.~Geiger, P.~Lenz, C.~Stiller, and R.~Urtasun, ``Vision meets robotics: The kitti dataset,'' \emph{International Journal of Robotics Research (IJRR)}, vol.~32, 2013. [Online]. Available: \url{http://www.cvlibs.net/datasets/kitti.}
\BIBentrySTDinterwordspacing

\bibitem{MoCo}
K.~He \emph{et~al.}, ``Momentum contrast for unsupervised visual representation learning,'' in \emph{2020 IEEE/CVF Conference on Computer Vision and Pattern Recognition (CVPR)}, 2020, pp. 9729--9738.

\bibitem{misra2020self}
I.~Misra and L.~v.~d. Maaten, ``Self-supervised learning of pretext-invariant representations,'' in \emph{2020 IEEE/CVF Conference on Computer Vision and Pattern Recognition (CVPR)}, 2020, pp. 6707--6717.

\bibitem{chen2021empirical}
X.~Chen, S.~Xie, and K.~He, ``An empirical study of training self-supervised vision transformers,'' in \emph{ICCV}, 2021, pp. 9640--9649.

\bibitem{hadsell2006dimensionality}
R.~Hadsell, S.~Chopra, and Y.~LeCun, ``Dimensionality reduction by learning an invariant mapping,'' in \emph{2006 IEEE/CVF Conference on Computer Vision and Pattern Recognition (CVPR)}, vol.~2.\hskip 1em plus 0.5em minus 0.4em\relax IEEE, 2006, pp. 1735--1742.

\bibitem{SimCLR}
T.~Chen \emph{et~al.}, ``A simple framework for contrastive learning of visual representations,'' in \emph{International conference on machine learning}.\hskip 1em plus 0.5em minus 0.4em\relax PMLR, 2020, pp. 1597--1607.

\bibitem{Fuerst2024}
\BIBentryALTinterwordspacing
M.~Fürst \emph{et~al.}, ``Learned fusion: 3d object detection using calibration-free transformer feature fusion,'' in \emph{International Conference on Pattern Recognition Applications and Methods (ICPRAM-2024)}, 2024. [Online]. Available: \url{https://orcid.org/0009-0000-0711-229X}
\BIBentrySTDinterwordspacing

\bibitem{cka-kornblith2019similarity}
S.~Kornblith \emph{et~al.}, ``Similarity of neural network representations revisited,'' in \emph{International conference on machine learning}.\hskip 1em plus 0.5em minus 0.4em\relax PMLR, 2019, pp. 3519--3529.

\bibitem{cka-cortes2012algorithms}
C.~Cortes, M.~Mohri, and A.~Rostamizadeh, ``Algorithms for learning kernels based on centered alignment,'' \emph{The Journal of Machine Learning Research}, vol.~13, pp. 795--828, 2012.

\bibitem{raghu2021vision}
M.~Raghu \emph{et~al.}, ``Do vision transformers see like convolutional neural networks?'' \emph{Advances in neural information processing systems}, vol.~34, pp. 12\,116--12\,128, 2021.

\end{thebibliography}
\bibliographystyle{IEEEtran}

\end{document}